\documentclass[letterpaper, 10 pt, conference]{ieeeconf}  % Comment this line out if you need a4paper

\IEEEoverridecommandlockouts                              % This command is only needed if 
\usepackage{graphics} % for pdf, bitmapped graphics files
\usepackage{epsfig} % for postscript graphics files
\usepackage{mathptmx} % assumes new font selection scheme installed
\usepackage{times} % assumes new font selection scheme installed
\usepackage{amsmath} % assumes amsmath package installed
\usepackage{amssymb}  % assumes amsmath package installed
\usepackage{algorithm}  
\usepackage{algorithmicx}  
\usepackage{algpseudocode}  
\usepackage{multirow}
\usepackage{caption}
\usepackage{cite}
\usepackage[pagebackref=true,breaklinks=true,colorlinks,bookmarks=false]{hyperref}

\newcommand{\etal}{\emph{et al.}}
\newcommand{\eg}{\emph{e.g. }}
\newcommand{\ie}{\emph{i.e. }}

\title{\LARGE \bf
FGR: Frustum-Aware Geometric Reasoning for Weakly Supervised 3D Vehicle Detection
}

\author{Yi Wei, Shang Su, Jiwen Lu, and Jie Zhou% <-this % stops a space
	\thanks{Yi Wei, Shang Su, and Jiwen Lu are with the State Key Lab of Intelligent Technologies and Systems, Beijing National Research Center for Information Science and Technology (BNRist), and the Department of Automation, Tsinghua University, Beijing, 100084, China. Email: y-wei19@mails.tsinghua.edu.cn; sus17@mails.tsinghua.edu.cn; lujiwen@tsinghua.edu.cn. Jiwen Lu is the corresponding author of this paper. Jie Zhou is with the State Key Lab of Intelligent Technologies and Systems, Beijing National Research Center for Information Science and Technology (BNRist), the Department of Automation, Tsinghua University, and the Tsinghua Shenzhen International Graduate School, Tsinghua University, Shenzhen 518055, China. Email: jzhou@tsinghua.edu.cn.}}

\begin{document}

\maketitle
\thispagestyle{empty}
\pagestyle{empty}

%%%%%%%%%%%%%%%%%%%%%%%%%%%%%%%%%%%%%%%%%%%%%%%%%%%%%%%%%%%%%%%%%%%%%%%%%%%%%%%%
\begin{abstract}

In this paper, we investigate the problem of weakly supervised 3D vehicle detection. Conventional methods for 3D object detection need vast amounts of manually labelled 3D data as supervision signals. However, annotating large datasets requires huge human efforts, especially for 3D area. To tackle this problem, we propose frustum-aware geometric reasoning (FGR) to detect vehicles in point clouds without any 3D annotations. Our method consists of two stages: coarse 3D segmentation and 3D bounding box estimation. For the first stage, a context-aware adaptive region growing algorithm is designed to segment objects based on 2D bounding boxes. Leveraging predicted segmentation masks, we develop an anti-noise approach to estimate 3D bounding boxes in the second stage. Finally 3D pseudo labels generated by our method are utilized to train a 3D detector. Independent of any 3D groundtruth, FGR reaches comparable performance  with fully supervised methods on the KITTI dataset. The findings indicate that it is able to accurately detect objects in 3D space with only 2D bounding boxes and sparse point clouds. 

\end{abstract}

%%%%%%%%%%%%%%%%%%%%%%%%%%%%%%%%%%%%%%%%%%%%%%%%%%%%%%%%%%%%%%%%%%%%%%%%%%%%%%%%
\section{INTRODUCTION}
In recent years, great achievements have been made in 3D area \cite{pon2020object,wang2020pointtracknet,battrawy2019lidar,wei2019conditional,qi2017pointnet++,griffiths2020finding}. Beyond 2D perception \cite{deng2009imagenet,simonyan2014very,huang2017densely,ren2015faster,liu2016ssd,he2017mask,li2017fully,wei2018quantization}, 3D scene understanding is indispensable and important because of wide real-world appliations. As one of hot 3D topics , 3D object detection \cite{qin2019monogrnet, li2019stereo, li2019gs3d,srivastava2019learning,simonelli2019disentangling,wang2019pseudo,ku2019improving,brazil2019m3d,liu2019deep,qi2019deep} focuses on the problem of detecting objects' tight bounding boxes in 3D space. It has attracted more and more attention due to eager demand in computer vision and robotics, \eg  autonomous driving, robot navigation and augmented reality. Point clouds are one of the key 3D representations, which consist of points in 3D space. Since point clouds provide rich geometric information, many 3D object detection methods \cite{ku2018joint, yi2020segvoxelnet, zhou2019iou, chen2019fast, yan2018second, liang2019multi,shi2019pv}  use them as inputs.  

Thanks to the development of deep learning, many approaches \cite{ mousavian20173d,yan2018second, liang2019multi,shi2019pv} adopt neural networks for 3D object detection, which achieve remarkable performances on various benchmarks \cite{dai2017scannet,geiger2012we,song2015sun}. Despite high accuracy, these learning-based methods require large amounts of 3D groundtruth. However, annotating these 3D data is a complex task demanding accurate quality control. The labeling process is expensive and time-consuming even utilizing crowd-sourcing systems (\eg MTurk).  Compared with 3D labels, Tang \etal \cite{tang2019transferable} mention that labeling 2D bounding boxes can be 3-16 times faster. Benefitting from large-scale 2D datasets, such as ImageNet \cite{deng2009imagenet} and MS COCO \cite{lin2014microsoft}, we can also obtain accurate 2D bounding boxes from high-performance 2D detectors. In addition, with the improvement of 3D sensors, it becomes much easier to get high-quality raw 3D training data. Thus, a valuable direction is exploring how to leverage 2D bounding boxes along with sparse point clouds from LiDAR for 3D object detection. 

\begin{figure}[tb]
	\centering
	\includegraphics[width=0.85\linewidth]{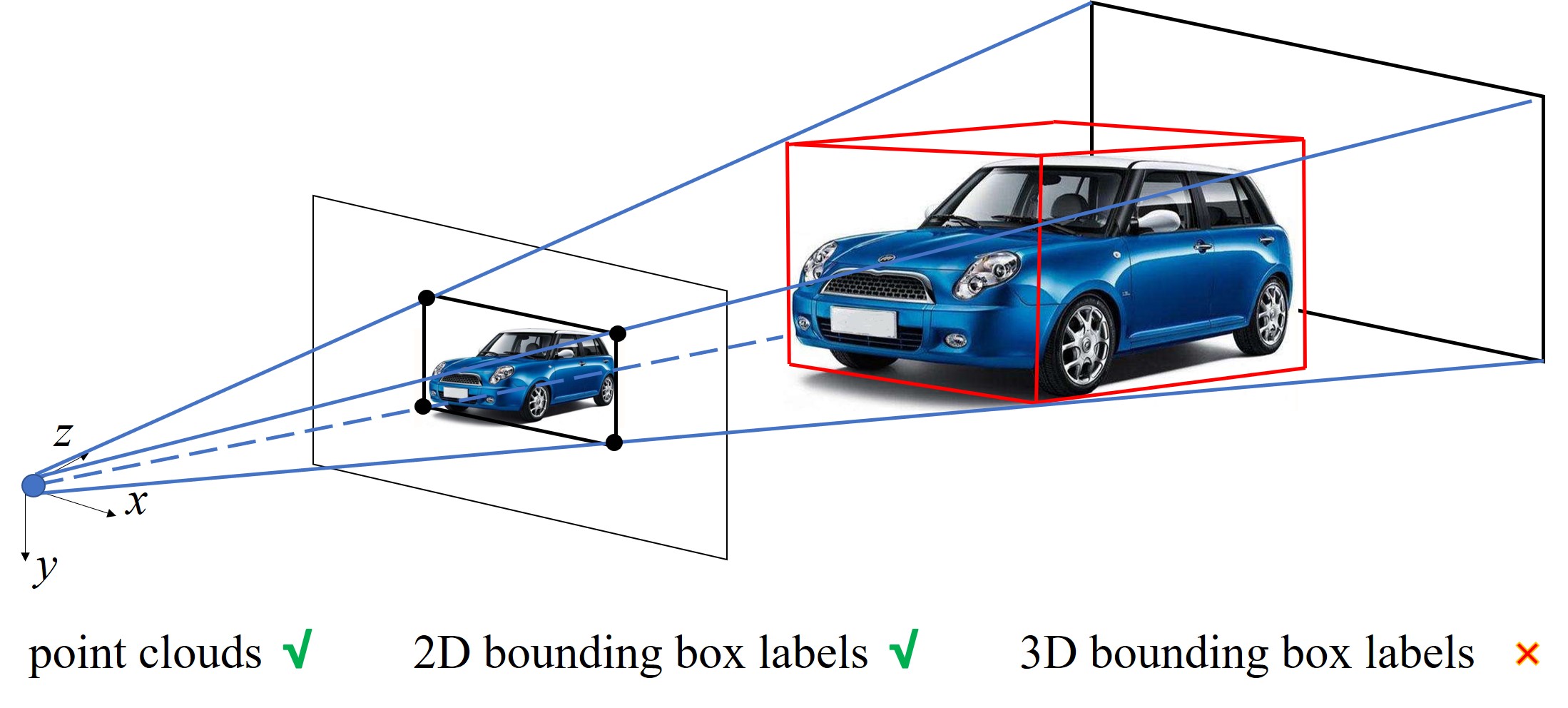}
	\caption{An overview of our proposed framework. In our algorithm, we only need 2D labels along with sparse point clouds from LiDAR. We find that semantic information in 2D bounding boxes and 3D structure information in point clouds are enough for 3D vehicle detection.}
	\label{fig:overview}
	\vspace{-6mm}
\end{figure}

 In this work, we propose frustum-aware geometric reasoning (FGR) for weakly supervised 3D vehicle detection. Fig. \ref{fig:overview} illustrates the overview of our method. Our framework only leverages 2D bounding boxes annotations and sparse point clouds. It consists of two components: coarse 3D segmentation and 3D bounding box estimation. In coarse 3D segmentation stage, we first employ RANdom SAmple Consensus (RANSAC) \cite{fischler1981random} to remove the ground part and extract the frustum area for each object. To tackle cases with different point densities, we propose a context-aware adaptive region growing method that leverages contextual information and adjusts the threshold adaptively. The aim of 3D bounding box estimation stage is to calculate 3D bounding boxes of segmented vehicle point clouds.  More specifically, we design an anti-noise framework to estimate bounding rectangles in Bird's Eye View (BEV) and localize the key vertex.  Then, the key vertex along with its edges are used to intersect the frustum. Based on 2D bounding boxes, our pipeline can generate pseudo 3D labels , which are finally used to train a 3D detector. To evaluate our method, we conducted experiments on KITTI Object Detection Benchmark \cite{geiger2012we}. Experimental results show that FGR achieves comparable performance with fully supervised methods and can be applied as a 3D annotator given 2D labels. The code is available at {\small \url{https://github.com/weiyithu/FGR}}.  
 
\section{Related Work}
\noindent \textbf{LiDAR-based 3D Object Detection:} 
Obtaining high-quality 3D data, LiDAR-based methods \cite{ku2018joint, yi2020segvoxelnet, zhou2019iou, chen2019fast, yan2018second, liang2019multi, shi2019pv}  are able to predict more accurate results than image-based methods. AVOD \cite{ku2018joint} adopts RPN to generate 3D proposals, which leverages a detection network to perform accurate oriented 3D bounding box regression and category classification. Lang \etal \cite{lang2019pointpillars} develop a well-designed encoder called PointPillars to learn the representation of point clouds organized in vertical columns. Similar to PointPillars, PointRCNN \cite{shi2019pointrcnn} also utilizes PointNet \cite{qi2017pointnet} as the backbone. It consists of two stages and directly uses point clouds to produce 3D proposals. Following \cite{shi2019pointrcnn}, to better refine proposals, STD \cite{yang2019std} designs a sparse to dense strategy.  Beyond PointRCNN \cite{shi2019pointrcnn}, Part-$A^2$ \cite{shi2020points} has the part-aware stage and the part-aggregation stage, which can better leverages free-of-charge part supervisions from 3D labels. Compared with two-stage detectors, single-stage detectors \cite{yang20203dssd, he2020structure} are more efficient. 3DSSD \cite{yang20203dssd} adopts a delicate box prediction network including a candidate generation layer and an anchor-free regression head to balance speed and accuracy. SA-SSD \cite{he2020structure} leverages the structure information of 3D point clouds with an auxiliary network .

\noindent \textbf{Frustum-based 3D Object Detection:} 
Some of 3D object detection methods \cite{shen2019frustum,qi2018frustum,wang2019frustum} first extract frustum area which is tightly fitted with 2D bounding boxes. In particular, F-PointNet \cite{qi2018frustum} predicts 3D bounding boxes from the points in frustums and achieves efficiency as well as high recall for small objects. It can also handle strong occlusion or cases with very sparse points. Beyond F-PointNet, F-ConvNet \cite{wang2019frustum} aggregates point-wise features as frustum level feature vectors and arrays these aggregated  features as the feature map. It also applies FCN to fuse frustum-level features and refine features over a reduced 3D space. Frustum VoxNet \cite{shen2019frustum} is another work relying on frustums. Instead of using the whole frustums, it determines the most interested parts and voxelize these parts of the frustums. 
Inspired by these methods, we also utilize frustum area to calculate 3D bounding boxes.

\noindent \textbf{Weakly Supervised / Semi-supervised 3D Object Detection:} 
Seldom works \cite{tang2019transferable, qin2020weakly, meng2020weakly, zhao2020sess} focus on weakly supervised or semi-supervised 3D object detection.  Leveraging 2D labels and part of 3D labels, Tang \etal \cite{tang2019transferable} propose a cross-category semi-supervised method (CC), which transfers 3D information from other classes to the current class while maintaining 2D-3D consistency. With weaker supervision, our method still outperforms CC for a large margin. Also without using any 3D labels, VS3D \cite{qin2020weakly} only adopts a pretrained model as the supervision. However, the performance of their method is much worse than fully-supervised methods. Recently, Meng \etal \cite{meng2020weakly} propose a two-stage architecture for weakly supervised 3D object detection, which requires a few precisely labeled object instances. Compared with their method, our FGR does not need any 3D labels. Mentioned in \cite{tang2019transferable, qin2020weakly}, annotating 3D bounding boxes is time-consuming and it's worthy exploring 3D object detection from only 2D labels. 

\begin{figure*}[tb]
	\centering
	\includegraphics[width=0.95\linewidth]{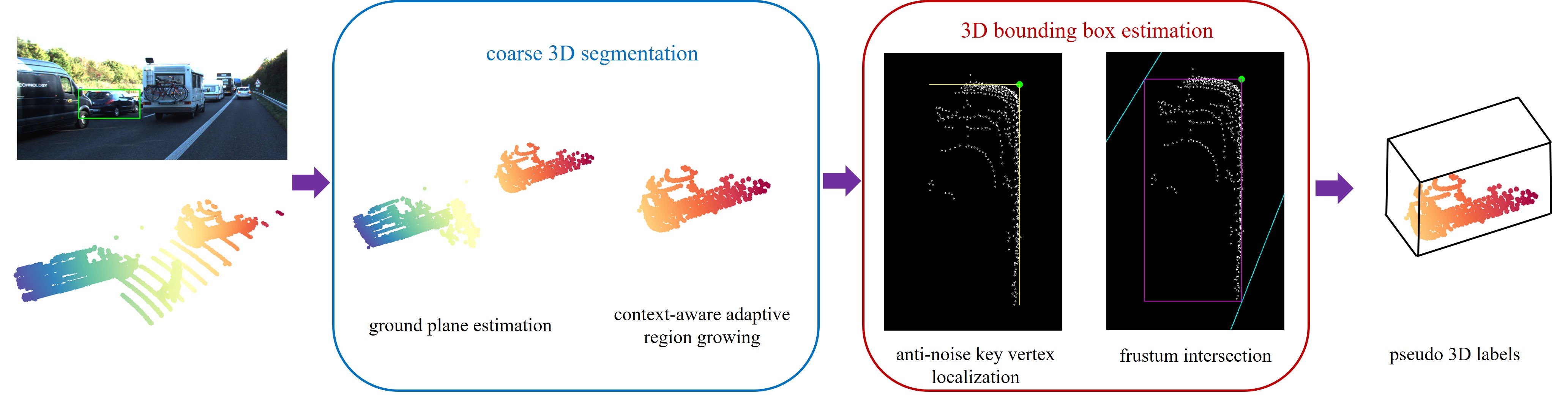}
	\caption{The pipeline of FGR. Our framework consists of two parts. For coarse 3D segmentation, we first estimate ground plane and remove it from the whole point clouds. Then we adopt context-aware adaptive region growing algorithm to get segmentation mask. However, there still exists noise in segmented points set. To solve this problem, we propose the anti-noise key vertex localization. Finally we use key vertex and key edges to intersect the frustum to predict 3D bounding boxes.}
	\label{fig:pipeline}
	\vspace{-1mm}
\end{figure*}

\section{Frustum-aware Geometric Reasoning}
Given 2D bounding boxes and LiDAR point clouds, our method aims at 3D vehicle detection without 3D labels. Fig. \ref{fig:pipeline} illustrates the pipeline of frustum-aware geometric reasoning. Our framework can generate pseudo 3D labels, which is then used to train a 3D detector. 
There are two components in our approach: coarse 3D segmentation and 3D bounding box estimation. We will introduce these two parts respectively in this section.

\subsection{Coarse 3D Segmentation}

\renewcommand{\algorithmicrequire}{\textbf{Input:}}  
\renewcommand{\algorithmicensure}{\textbf{Output:}}

A straightforward method for point clouds segmentation is applying region growing algorithm \cite{adams1994seeded} to separate out objects from frustum area, which is regarded as our baseline. However, our experimental results demonstrate that the baseline method is not accurate. 

\noindent \textbf{Ground Plane Estimation:} 
We observe that because most of vehicles lying on the ground, ground points will heavily affect region growing results. To alleviate the influence, we first adopt RANSAC \cite{fischler1981random} to estimate ground plane and remove it.

\noindent \textbf{Context-aware Adaptive Region Growing:} 
It is difficult to detect the vehicle which is highly occluded by others. However, if we can first detect other vehicles and remove their points from the whole point clouds, it will be much easier to segment current vehicle. Given a series of 2D bounding boxes $\{B_i\}$, we extract frustums and sort them according to the median depth of points in frustums. The vehicles nearer to camera will be executed with algorithm earlier.
\begin{algorithm*}[tb]
	\setlength{\belowcaptionskip}{-20pt}
	\caption{ The pipeline of context-aware adaptive region growing}
	\label{alg}  
	\begin{algorithmic}[1] 
		\Require Whole point cloud $P_{all}$,  \ points set in frustum region $F^i$ of vehicle $O^i$, \ distance thresholds $\{\phi^k\}$
		\Ensure Segmented points set $M_{best}^i$
		\For{$k=1$ to $n$:}
		\While{ there exists a point $P^i_j$ in $F_i$ which doesn't  belong to any connected component in $\{C_{jk}^{i}\}$}
		\State Set up an empty points set  $C_{jk}^{i}$, add $P^i_j$ to $C_{jk}^{i}$
		\While{there exists a point $p$ in $C_{jk}^{i}$ which has not been processed}
		\State Search points set $q$ from whole point cloud $P_{all}$ whose Euclidean distance to $p$ is smaller than $\phi^k$
		\State For each point in $q$, \ check if the point has already been processed, \  and if not add it to $C_{jk}^{i}$
		\EndWhile
		\State If $\frac{|C_{jk}^{i}  \cap F^i|}{|C_{jk}^{i}|} < \theta_{seg}$, we treat this connected component as a outlier and remove it
		\EndWhile
		\State From connected component set $\{C_{jk}^i\}$, select $M^i_k$ which has the most points as the segmentation result for  $\phi^k$
		\EndFor
		\State From $\{M^i_k\}$, select $M_{best}^i$ which has the most points as the final segmentation result.
		
	\end{algorithmic} 
\end{algorithm*} 

Algorithm \ref{alg} shows the pipeline of context-aware adaptive region growing. For a distance threshold $\phi_{k}$, we treat two points as neighbour points if their Euclidean distance is smaller than $\phi_k$. According to this definition, for each vehicle $O^i$, we randomly select a point $P_{j}^i$ in frustum area and  calculate its connected component $C_{jk}^{i}$. We repeat the previous steps until each point in $F^i$ belongs to a connected component. Finally, the vehicle segmentation mask is selected from  $\{C_{jk}^{i}\}$. Among these connected components, we filter out $C_{jk}^{i}$ if it satisfies following criterion:
\begin{equation}
\frac{|C_{jk}^{i}  \cap F^i|}{|C_{jk}^{i}|} < \theta_{seg}
\end{equation} 
where $\theta_{seg}$ is a threshold and $| \cdot |$ means the size of the set. This criterion indicates that if the proportion of a connected component inside frustum is smaller than a threshold, we consider this component belongs to other objects but not current vehicle. After removing these outliers, we choose the connected component $M^i_k$ which has the most points as the segmentation set of vehicle $O^i$.

For each threshold $\phi_k$, we calculate segmentation set $M^i_k$. It's reasonable to set different $\phi_k$ for different objects according their point densities. Thus, we traverse threshold set $\{\phi_k\}$ and select the segmented points set which has the most points as the final segmentation result $M_{best}^i$.

\subsection{3D Bounding Box Estimation}
A baseline method to calculate 3D bounding boxes of segmented points set $M_{best}^i$ is to estimate minimum-area rectangle which encases all points. Although this is a mature algorithm, it is sensitive to noise points and doesn't make use of frustum information. Different from this baseline approach, for the vehicle $O^i$, we first locate the key vertex $V^i$ of $M_{best}^i$ in an anti-noise manner. Then based on this key vertex, we calculate the final 3D bounding box by computing the intersection with frustum boundaries. This stage is conducted on Bird's Eye View (BEV).

\begin{figure}[tb]
	\centering
	\includegraphics[width=0.9\linewidth]{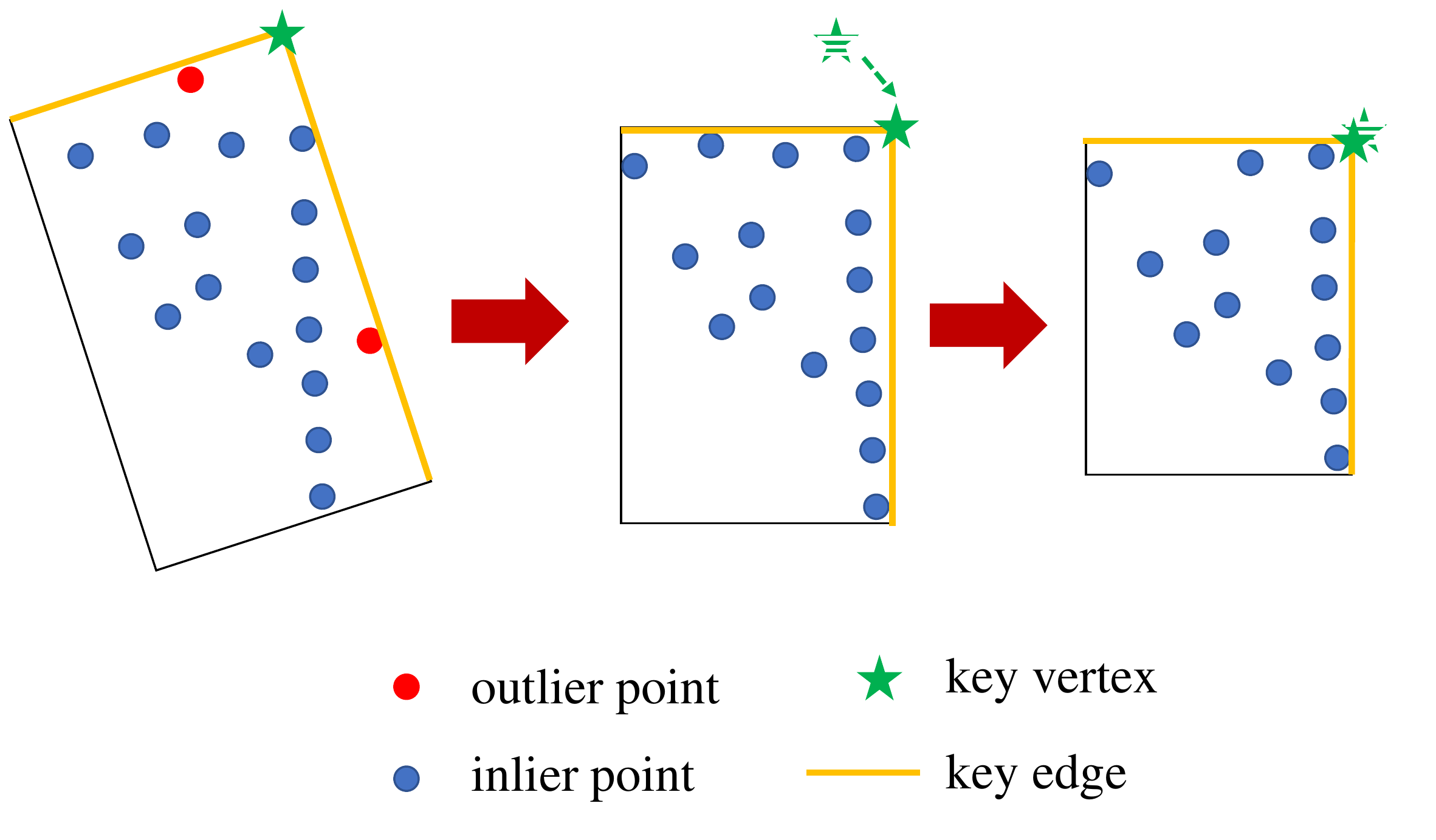}
	\caption{The illustration of anti-noise key vertex localization. Adopting a iterative framework, our method can detect noise points and remove them. The algorithm will stop if key vertex's position changes slightly between two iterations.}
	\label{fig:vertex}
	\vspace{-6mm}
\end{figure}

\noindent \textbf{Anti-noise Key Vertex Localization:} 
Fig. \ref{fig:vertex} illustrates the procedure of anti-noise key vertex localization. For a bounding rectangle, we can separate out four right triangles according to its four vertexes. We define the right angle vertex of the triangle containing the most points as the key vertex and its corresponding legs as key edges. 

\begin{figure}[tb]
	\centering
	\includegraphics[width=0.7\linewidth]{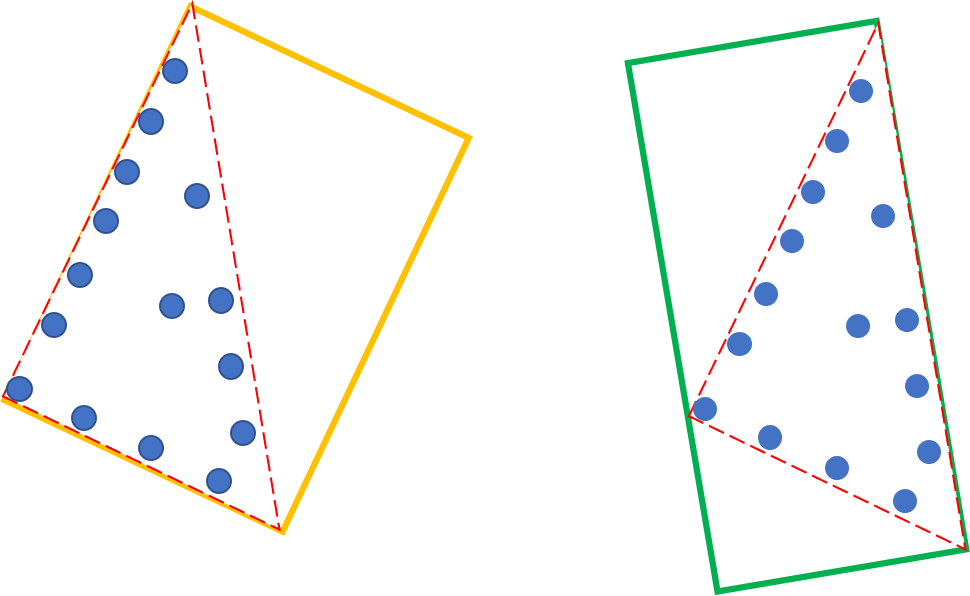}
	\caption{Using the area as the objective function will bring ambiguity. For the points in the red triangle, there are two optimal bounding rectangles (yellow rectangle and green rectangle have the same area). However, the yellow rectangle is what we want. }
	\label{fig:ambiguity}
	\vspace{-2mm}
\end{figure}

In many cases, we can only observe points in one right triangle. If we use the area as the objective function to calculate the optimal bounding rectangle, we will get two results, which are shown in Fig. \ref{fig:ambiguity}. To address this problem, we design a new objective function. We assume that we want to estimate the bounding rectangle of points set $Q$, and key edges are $l_1, l_2$ respectively. The following equation describes our new objective equation $f$:

\begin{equation}
\begin{aligned}
f &= \frac{|Q_{0}|}{|Q|} \\
Q_{0} = \{q|\ q \in Q, &\ ||q, l_1|| > \theta_{rect} , \ ||q, l_2|| > \theta_{rect}\}
\end{aligned}
\end{equation} 
where $||q, l||$ is the vertical distance between point $q$ and edge $l$, and $\theta_{rect}$ is the threshold. If most points are close to key edges, we think this bounding rectangle is well-estimated. We first enumerate rectangles encasing $Q$ whose orientations are from 0 degree to 90 degrees with 0.5 degree interval. Then we select one which has minimum $f$ as the bounding rectangle of $Q$.

Because accurate segmentation labels are unavailable, our predicted segmentation masks always obtain noise, which will heavily influence the calculation of bounding rectangles. For the noise point, if we delete it from $Q$ and recompute bounding rectangle, the position of the key vertex will vary a lot. In contrast, the key vertex is stable if we remove a few inlier points because there are many points lying on the key edges. In light of this discovery, we develop an iterative framework. For an estimated bounding rectangle, we delete points near to the key edges and recompute a new bounding rectangle. If the key vertex changes slightly, we treat the key vertex of this new bounding rectangle as $V^i$, and if not we will repeat above process until satisfying this condition. 

\begin{figure}[tb]
	\centering
	\includegraphics[width=0.8\linewidth]{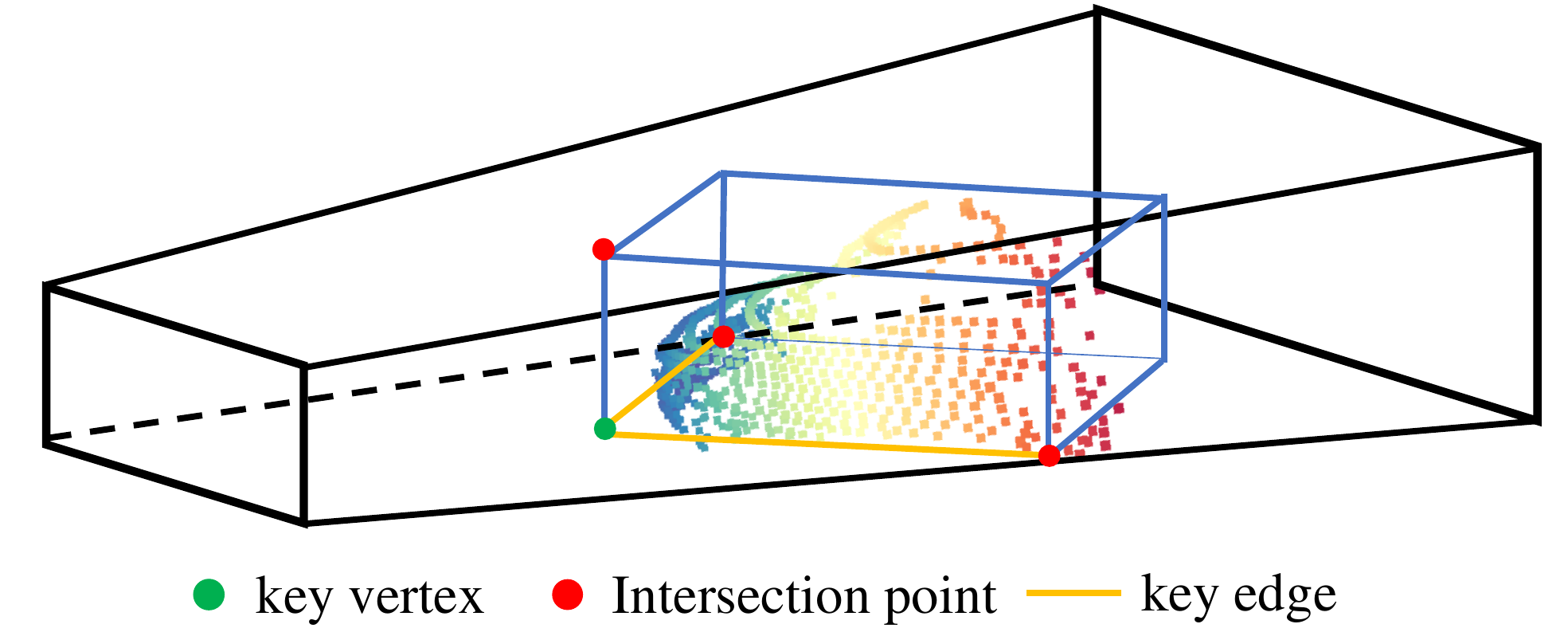}
	\caption{We use the key vertex and key edges to intersect the frustum. From these intersection points, we can get final 3D bounding box. }
	\label{fig:intersection}
	\vspace{-5mm}
\end{figure}

\noindent \textbf{Frustum Intersection:} 
Due to the occlusion, the observed points sometimes only cover a part of the whole vehicle. In other words, it's inaccurate to directly use bounding rectangles as the final prediction. Fortunately, we are able to compute accurate results leveraging 2D bounding boxes (frustums). From Fig. \ref{fig:intersection}, we can easily get 8 vertexes of the 3D bounding boxes given the key vertex and key edges of bounding rectangles. 

\section{Experiments}
In this section, we conduct extensive experiments to verify the effectiveness of our method. First we will describe the dataset and implementation details. Then we will exhibit main results on KITTI dataset \cite{geiger2012we}. Experiments in \S\ref{sec:ablation} verify the rationality of our method. Finally, we will show some qualitative results. 

\subsection{Dataset and Implementation Details}
\noindent \textbf{KITTI Dataset:} 
As a frequently used 3D object detection benchmark, the KITTI dataset  \cite{geiger2012we} contains 7,481 training pairs and 7,518 testing pairs of RGB images and point clouds. We follow \cite{chen2017multi} to divide original training samples into train split (3712 samples) and val split (3769 samples). Our algorithm was evaluated on car category samples. For experiments using 3D detectors, we report Average Precision for 3D and  Bird's Eye View (BEV) tasks. According to the 2D bounding box height, occlusion and truncation level, the benchmark is divided into three difficulty levels: Easy, Moderate and Hard. When evaluating the quality of generated labels, we set mean IoU and precision as evaluation metric.  To better evaluate precision, we utilized IoU threshold of 0.3, 0.5 and 0.7 respectively.

\noindent \textbf{Implementation Details:} 
In KITTI dataset \cite{geiger2012we}, vehicles' roll/pitch angle can be seemed as zero and we only need to estimate yaw angle (\ie orientation). Therefore, we can leverage Bird's Eye View (BEV) to detect vehicles instead of operating on 3D space. We set threshold $\theta_{seg}$ as 0.8  for context-aware adaptive region growing. $\{\phi^k\}$ was enumerated from 0.1 to 0.7 with 0.1 interval. As for anti-noise key vertex localization,  we choosed one-tenth length of key edges as $\theta_{rect}$. During the localization process, if the change of key vertex's position was smaller than 0.01 meter, we would stop iterative algorithm and get final bounding rectangles.  PointRCNN \cite{shi2019pointrcnn} was adopted as our 3D detector backbone. We finally filtered 3D bounding boxes with extremely wrong sizes.

\begin{table*}[t!]
	\centering
	\caption{Performance comparison with other fully supervised methods on KITTI test set.}
	\resizebox{0.7\textwidth}{!}{
		\begin{tabular}{|c|c||ccc|ccc|}
			\hline
			\multirow{2}{*}{Method}  & \multirow{2}{*}{3D labels}& \multicolumn{3}{c|}{AP\textsubscript{3D}(IoU=0.7)} & \multicolumn{3}{c|}{AP\textsubscript{BEV}(IoU=0.7)}   \\   
			 & & \multicolumn{1}{c}{Easy} & \multicolumn{1}{c}{Moderate} & Hard  
			 & \multicolumn{1}{c}{Easy} & \multicolumn{1}{c}{Moderate} & Hard     \\   \hline
			MV3D \cite{chen2017multi}& \checkmark &74.97& 63.63 &	54.00 &86.62 & 78.93 &69.80  \\
			AVOD \cite{ku2018joint}& \checkmark & 76.39 &66.47 & 60.23 &89.75& 84.95&	78.32  \\
			F-PointNet \cite{qi2018frustum} & \checkmark&  82.19 &69.79 &60.59 &91.17 &84.67 &74.77  \\
			SECOND \cite{yan2018second}& \checkmark&  83.34 &72.55 &65.82&89.39 &83.77 &78.59  \\
            PointPillars \cite{lang2019pointpillars} & \checkmark&  82.58 &74.31 &68.99&90.07 &86.56 &82.81  \\
			PointRCNN \cite{shi2019pointrcnn}& \checkmark&  86.96 &75.64 &70.70 & 92.13& 87.39& 82.72   \\
			SegVoxelNet \cite{yi2020segvoxelnet}& \checkmark&  86.04 &76.13 &70.76 & 91.62& 86.37& 83.04   \\
			F-ConvNet \cite{wang2019frustum} & \checkmark&87.36 &76.39 &66.69& 91.51 & 85.84 &76.11 \\
			Part-$A^2$ \cite{shi2020points} & \checkmark&87.81 &78.49&	73.51 &	91.70 & 87.79 & 84.61  \\
			SASSD \cite{he2020structure} & \checkmark&88.75 &79.79 &74.16  & 95.03 & 91.03 &85.96  \\ \hline
			%PV-RCNN \cite{shi2019pv} & \checkmark&92.57 &84.83 &82.69 & 94.98 &90.65 &86.14 \\ \hline
			FGR  & $\times$& 80.26& 68.47& 61.57& 90.64& 82.67& 75.46\\
			 \hline
	\end{tabular}}

	\vspace{-4mm}
	\label{tab:test}
\end{table*}

\begin{table}[tb]
	\centering
	\caption{Performance comparison with other weakly supervised methods on KITTI val set.}
	\resizebox{0.37\textwidth}{!}{
		\begin{tabular}{|c||c|c|c|}
			\hline
            \multirow{2}{*}{Method} &  \multicolumn{3}{c|}{AP\textsubscript{3D}(Easy)}        \\   
			   &  \multicolumn{1}{c}{IoU=0.25} &\multicolumn{1}{c}{IoU=0.5} &\multicolumn{1}{c|}{IoU=0.7 }     \\   \hline 
            CC \cite{tang2019transferable} &69.78 &- &- \\
		    VS3D \cite{qin2020weakly}& - & 40.32& - \\
            WS3D \cite{meng2020weakly}& - &- & 84.04
			\\ \hline
			FGR & \bf{97.24} & \bf{97.08} & \bf{86.11}  \\
			\hline
	\end{tabular}}
	
	\label{tab:val}
\end{table}

\iffalse
\begin{table}[tb]
	\centering
	\caption{Performance comparison with other weakly supervised methods on KITTI val set.}
	\resizebox{0.4\textwidth}{!}{
		\begin{tabular}{|c|c||c|c|c|}
			\hline
			\multirow{2}{*}{IoU} & \multirow{2}{*}{Method} &  \multicolumn{3}{c|}{AP\textsubscript{3D}}        \\   
			 &    &  \multicolumn{1}{c}{Easy} &\multicolumn{1}{c}{Moderate} &\multicolumn{1}{c|}{Hard }     \\   \hline 
			 \multirow{2}{*}{0.25} &CC \cite{tang2019transferable} &69.78 &58.66 &51.40 \\
			&FGR  & \bf{97.24} & \bf{89.67} & \bf{89.06}  \\ \hline	
			\multirow{2}{*}{0.5}&VS3D \cite{qin2020weakly}& 40.32 & 37.36  & 31.09 \\
			&FGR &  \bf{97.08} & \bf{89.55} & \bf{88.76}  \\  \hline
			\multirow{2}{*}{0.7}&WS3D \cite{meng2020weakly}& 84.04 &75.10& \bf{73.29}
			\\
			 &FGR & \bf{86.11} & \bf{74.86} & 67.87  \\
			\hline
	\end{tabular}}
	
	\label{tab:val}
\end{table}
\fi

\begin{table}[tb]
	\centering
	\caption{Evaluation results with different 3D backbones on KITTI val set}
	\resizebox{0.45\textwidth}{!}{
		\begin{tabular}{|c|c||ccc|}
			\hline
			\multirow{2}{*}{Method} & \multirow{2}{*}{3D labels} &   \multicolumn{3}{c|}{AP\textsubscript{3D}}        \\   
			&    &  \multicolumn{1}{c}{Easy} &\multicolumn{1}{c}{Moderate} &\multicolumn{1}{c|}{Hard}     \\   \hline 
			PointRCNN \cite{shi2019pointrcnn} & \checkmark& 88.88 &78.63& 77.38
			 \\	
			F-ConvNet \cite{wang2019frustum} & \checkmark&  89.02 &78.80 &77.09  \\
			Part-$A^2$ \cite{shi2020points} & \checkmark&  89.47& 79.47 & 78.54    \\ \hline
			Ours (PointRCNN) & $\times$  &86.11 & 74.86 & 67.87  \\
			Ours (F-ConvNet) & $\times$  & 86.40 & 73.87 & 67.03  \\
			Ours (Part-$A^2$) & $\times$  & 86.33 & 73.75 & 67.53  \\
			\hline
	\end{tabular}}
	\vspace{-4mm}
	\label{tab:backbone}
\end{table}

\subsection{Main Results}  
Table \ref{tab:test} shows the comparison with other fully supervised methods on KITTI test set. Despite without using any 3D labels, our FGR still achieves comparable performance with some fully supervised methods. We also compare FGR with other weakly-supervised methods on KITTI validation set. However, the supervision of these methods are different with ours (\ie CC\cite{tang2019transferable} uses 2D labels and 3D labels from other classes, VS3D \cite{qin2020weakly} uses a 2D pretrained model and WS3D \cite{meng2020weakly} uses part of 3D labels) and we just report these results for reference.  From Table \ref{tab:val}, we can see that our approach outperforms these weakly supervised methods.  We should notice that despite the supervision of our method is weaker than CC\cite{tang2019transferable} , FGR still outperforms it for a large margin. To better demonstrate the effectiveness of FGR, we also show results using other 3d detector backbones in Table \ref{tab:backbone}. In addition, compared with fully supervised baselines, the performances of our FGR degrade marginally, which indicates that a potential application of FGR is to annotate 3D data given 2D bounding boxes and point clouds. These experimental results confirm that we are able to detect vehicles accurately without the aid of 3D groundtruth.

\subsection{Ablation Studies}  \label{sec:ablation}
We conducted experiments to confirm the effectiveness of each module in our method. To decouple the influence of 3D detectors, we evaluate the quality of generated labels in this subsection.
\begin{table}[tb]
		\vspace{-5mm}
	\centering
	\caption{Ablation studies for coarse 3D segmentation. }
	\resizebox{0.48\textwidth}{!}{
	\begin{tabular}{|c|c||c|c|c|c|}
		\hline
		ground plane & \multirow{2}{*}{adaptive} & \multirow{2}{*}{Mean IoU}  &   \multicolumn{3}{c|}{Precision}        \\   
		estimation&  &   &  \multicolumn{1}{c}{IoU=0.3} &\multicolumn{1}{c}{IoU=0.5} &\multicolumn{1}{c|}{IoU=0.7}     \\   \hline 
		& & 0.5311 & 74.40 & 61.05 & 41.09\\	
		$\checkmark$& & 0.6563 & 84.82 & 74.34 & 58.96 \\
		$\checkmark$&$\checkmark$ &  \bf{0.7845} & \bf{97.90} & \bf{96.70} & \bf{83.28}   \\ \hline
	\end{tabular}}
	
	\label{tab:ablation_seg}

\end{table}

\begin{figure}[tb]
	\centering
	\includegraphics[width=0.9\linewidth]{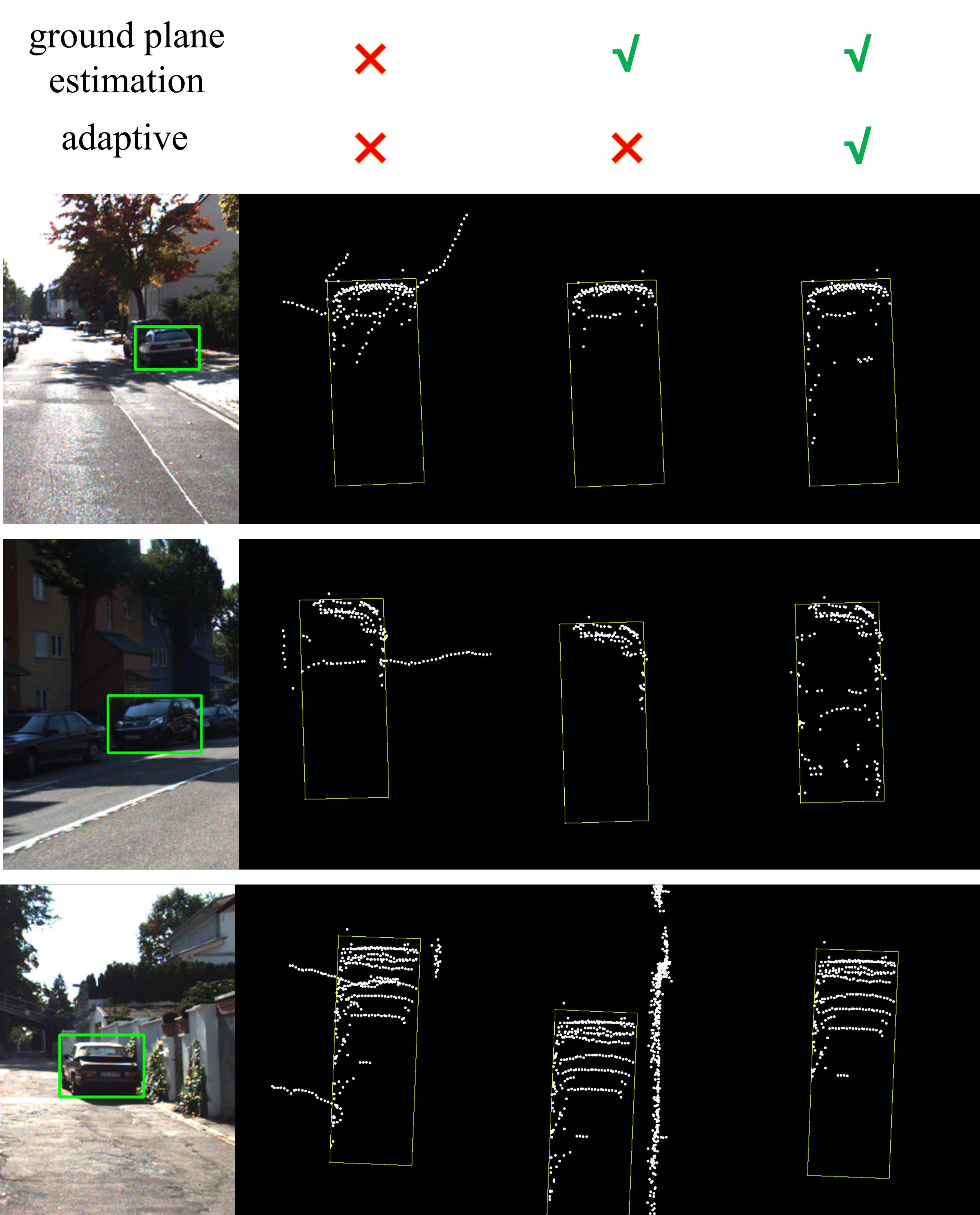}
	\caption{Ablation studies for coarse 3D segmentation. Without ground plane estimation, ground points will be segmented as inlier points. Also, the best threshold for different vehicles varys according to point densities. Yellow bounding boxes indicate groundtruth. Best viewed in BEV.  }
	\label{fig:ablation_seg}
	\vspace{-5mm}
\end{figure}

\noindent \textbf{Coarse 3D Segmentation:} 
We did ablation studies to verify the necessity of ground plane estimation and effectiveness of context-aware adaptive region growing. We fixed $\phi$ as 0.4 for the experiment that did not use adaptive thresholds. Table \ref{tab:ablation_seg} and Fig. \ref{fig:ablation_seg} illustrate quantitative and qualitative results respectively. From the second column in the figure, we can see that some ground points will be treated as inliers without ground plane estimation. In this way, the points of the vehicle will connect other regions through ground points, which accounts for a failed segmentation. From the third and fourth columns, we conclude that a fixed threshold $\phi$ is unable to handle all cases. On the one hand, if $\phi$  is too small, we will get incomplete segmentation masks (the first and second rows). On the other hand, if $\phi$  is too large, the segmented set will include many outliers (the third row).  The best threshold $\phi$ for these three cases are 0.7, 0.6 and 0.3 respectively.

\begin{figure*}[tb]
	\centering
	\includegraphics[width=1.0\linewidth]{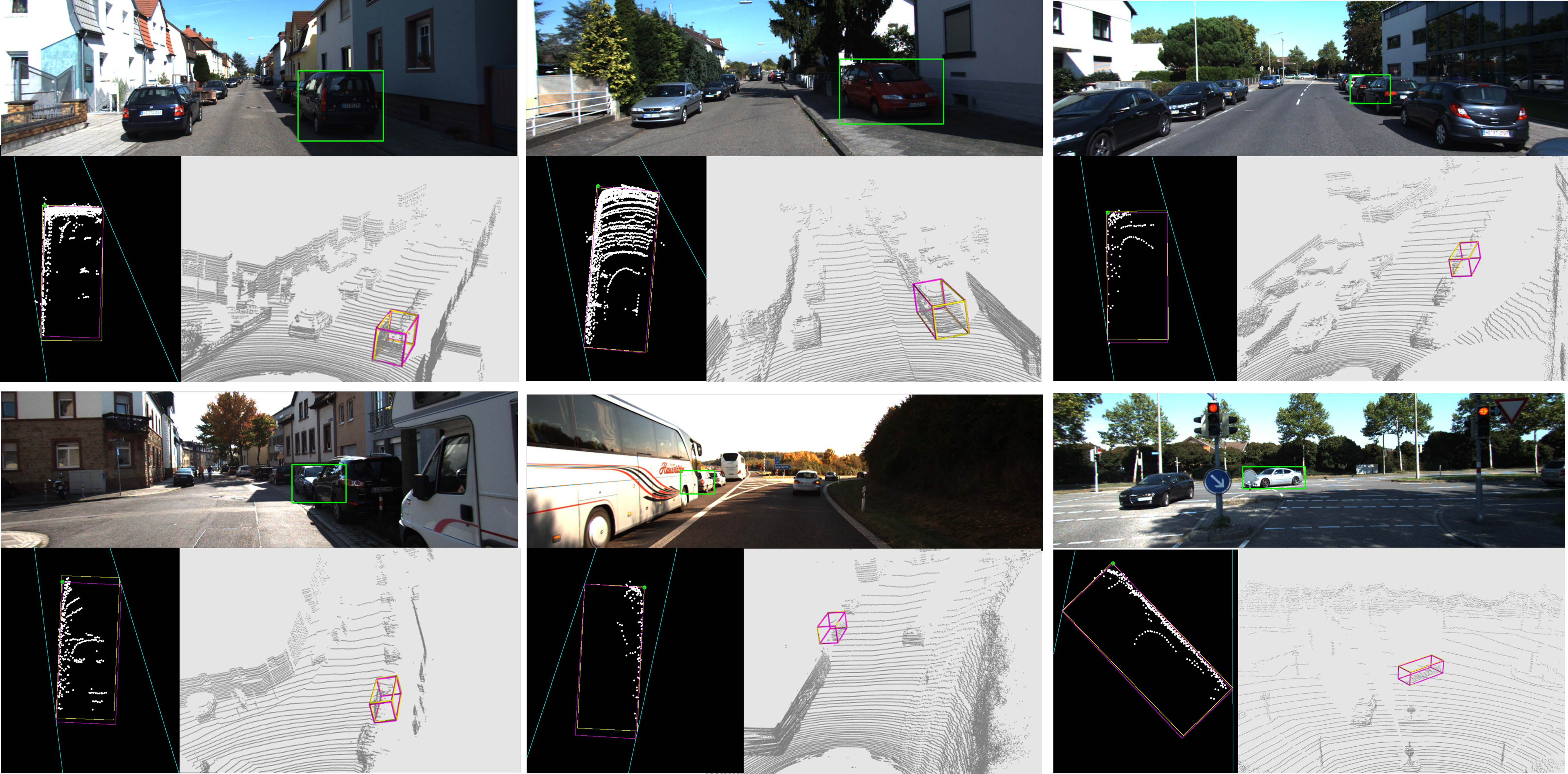}
	\caption{Qualitative results. We illustrate RGB images, Bird's Eye View (BEV) and LiDAR point clouds respectively. Predicted bounding boxes are drawn in purple while groundtruth is in yellow.}
	\label{fig:visual}
	\vspace{-3mm}
\end{figure*}
\begin{figure}
	\centering
	\includegraphics[width=0.9\linewidth]{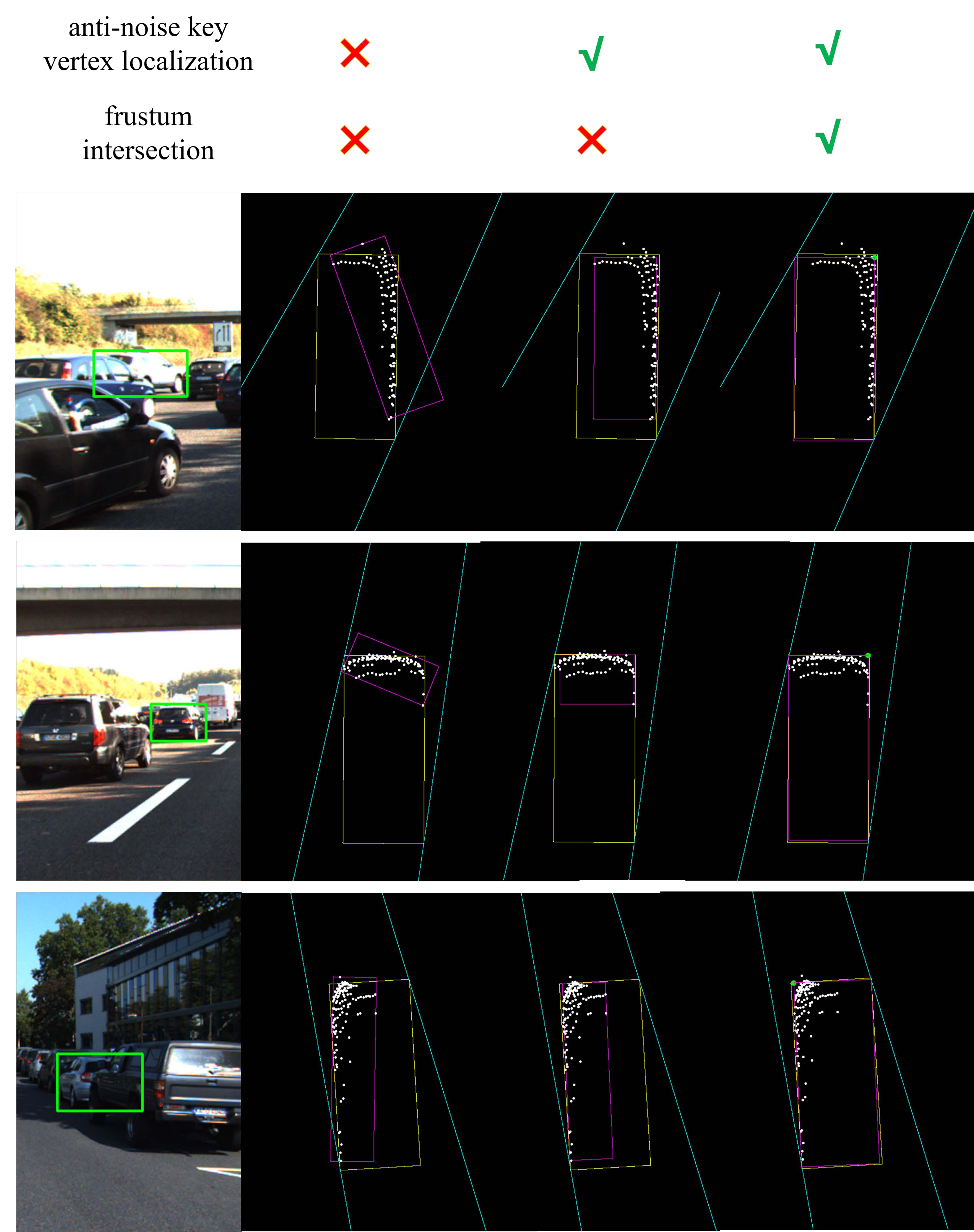}
	\caption{Ablation studies for 3D bounding box estimation. If we adopt baseline method,  the bounding rectangle will include noise points. Due to the occlusion, it's necessary to implement frustum intersection to better leverage 2D bounding boxes. Yellow bounding boxes indicate groundtruth while purple ones are prediction results. Green points represent key vertexes and cyan lines are frustum boundaries.  Best viewed in BEV.}
	\label{fig:ablation_rect}
	
\end{figure}
\begin{table}[tb]
	\centering
	\caption{Ablation studies for 3D bounding box estimation. }
	\resizebox{0.48\textwidth}{!}{
		\begin{tabular}{|c|c||c|c|c|c|}
			\hline
			anti-noise key & frustum & \multirow{2}{*}{Mean IoU}  &   \multicolumn{3}{c|}{Precision}        \\   
			vertex localization & intersection &   &  \multicolumn{1}{c}{IoU=0.3} &\multicolumn{1}{c}{IoU=0.5} &\multicolumn{1}{c|}{IoU=0.7}     \\   \hline 
			& & 0.5418 & 76.42 & 56.25 & 41.32\\	
			$\checkmark$& & 0.6255 & 89.37 & 69.31 & 54.50 \\
			$\checkmark$&$\checkmark$ &  \bf{0.7845} & \bf{97.90} & \bf{96.70} & \bf{83.28}  \\ \hline
	\end{tabular}}
	
	\label{tab:ablation_rect}
	\vspace{-3mm}
\end{table}
\noindent \textbf{3D Bounding Box Estimation:} 
In this part, we incrementally applied anti-noise key localization module and frustum intersection module to our framework. Table \ref{tab:ablation_rect} and Fig. \ref{fig:ablation_rect} exhibit experimental results. From the second column in the figure, we find that there still exists noise points after coarse 3D segmentation. If we directly operate on these noise points, we will get loose bounding rectangles. In addition, attributed to the occlusion, bounding rectangles tend to be smaller than groundtruth (the third column in Fig. \ref{fig:ablation_rect}). We will get wrong results if we directly use these bounding boxes. However, the position of the key vertex is correct, which can be leveraged to intersect frustums along with key edges.

\subsection{Qualitative Results}
To better illustrate the superiority of our method, we visualize some generated labels in Fig. \ref{fig:visual}. For simple cases which are non-occluded and have enough points, our method can achieve remarkable accuracy. Moreover, we are surprised to find that our method is also able to handle occlusion cases. These qualitative results demonstrate the effectiveness of FGR and also verify that it's possible to combine semantic information in 2D bounding boxes and 3D structure in point clouds for vehicle detection.

%\addtolength{\textheight}{-12cm}   % This command serves to balance the column lengths
                                  % on the last page of the document manually. It shortens
                                  % the textheight of the last page by a suitable amount.
                                  % This command does not take effect until the next page
                                  % so it should come on the page before the last. Make
                                  % sure that you do not shorten the textheight too much.

%%%%%%%%%%%%%%%%%%%%%%%%%%%%%%%%%%%%%%%%%%%%%%%%%%%%%%%%%%%%%%%%%%%%%%%%%%%%%%%%

%%%%%%%%%%%%%%%%%%%%%%%%%%%%%%%%%%%%%%%%%%%%%%%%%%%%%%%%%%%%%%%%%%%%%%%%%%%%%%%%

%%%%%%%%%%%%%%%%%%%%%%%%%%%%%%%%%%%%%%%%%%%%%%%%%%%%%%%%%%%%%%%%%%%%%%%%%%%%%%%%
%\section*{APPENDIX}

%Appendixes should appear before the acknowledgment.

%%%%%%%%%%%%%%%%%%%%%%%%%%%%%%%%%%%%%%%%%%%%%%%%%%%%%%%%%%%%%%%%%%%%%%%%%%%%%%%%

\section{Conclusion}
In this paper, we propose frustum-aware geometric reasoning for 3D vehicle detection without using 3D labels. The proposed framework consists of coarse 3D segmentation and 3D bounding box estimation modules. Our method achieves promising performance compared with fully supervised methods. Experimental results indicate that a potential application of FGR is to annotate 3D labels.  

\section*{ACKNOWLEDGMENT}

This work was supported in part by the National Natural Science Foundation of China under Grant U1713214, Grant U1813218, Grant 61822603, in part by Beijing Academy of Artificial Intelligence (BAAI), and in part by a grant from the Institute for Guo Qiang, Tsinghua University.

{\small
\bibliographystyle{IEEEtran}
\bibliography{IEEEexample}
}

\end{document}